\documentclass[runningheads]{llncs}
\usepackage{float}
\usepackage{graphicx}
\usepackage{cite}
\usepackage{amsmath,amssymb,amsfonts}
\usepackage{algorithmic}
\usepackage{graphicx}
\usepackage{textcomp}
\usepackage{xcolor}
\usepackage{todonotes}
\usepackage{caption}
\usepackage{epigraph}

\usepackage{adjustbox}
\usepackage{wrapfig}
\usepackage{makecell} 

\begin{document}

\title{Evaluating System Identification Methods for Predicting Thermal Dissipation of Heterogeneous SoCs\thanks{Part of this work was carried out with financial support from the Nordic Master programme  (contract NMP-2016/10169) and ECSEL-JU AIDOaRt project (grant agreement No 101007350).}}
%
%
\author{Joel Öhrling\orcidID{0000-0001-5259-8436} \and
Sébastien Lafond\orcidID{0000-0002-5286-5343} \and
Dragos Truscan\orcidID{0000-0002-4367-6225}}

\authorrunning{J. Öhrling et al.}
%
\institute{Åbo Akademi University, Turku, Finland\\
\email{joelohrling@gmail.com}, 
\email{\{sebastien.lafond,dragos.truscan@\}@abo.fi}}
\maketitle              
\begin{abstract}
In this paper we evaluate the use of system identification methods to build a thermal prediction model of heterogeneous SoC platforms that can be used to quickly predict the temperature of different configurations without the need of hardware. Specifically, we focus on modeling approaches that can predict the temperature based on the clock frequency and the utilization percentage of each core.  We investigate three methods with respect to their prediction accuracy: a linear state-space identification approach using polynomial regressors, a NARX neural network approach and a recurrent neural network approach configured in an FIR model structure. We evaluate the methods on an Odroid-XU4 board featuring an Exynos 5422 SoC. The results show that the model based on polynomial regressors significantly outperformed the other two models when trained with 1 hour and 6 hours of data. 
\end{abstract}

\section{Introduction}

In recent years, heterogeneous System-on-Chip (SoC) platforms have permeated many types of IT systems \cite{mcguinness_2013} \cite{ullman_2013} due to the efficient balance they provide between their computing power and power consumption. However, the challenge they provide is in choosing the correct configuration for a specific application workload. This is particularly difficult due to the very large configuration space.

Performing exhaustive testing on these types of systems becomes unfeasible, as the number of possible configurations is vast and requires the presence of hardware setups to experiment with. In order to speed up the exploration process, we investigate system identification methods to build a model of the platform that can be used to predict the temperature of different configurations quickly without the need of hardware. 

Modeling a processor based on theoretical relationship between the power and thermal dissipation requires extensive knowledge about the characteristics of the processor and its environment. For some processors, numerical values for the thermal characteristics of the materials and placement of the processor parts are readily available. However, for many processors, these values are not provided and have to be estimated or measured. Therefore, in this work, we focus on modeling approaches that can predict the temperature based on the clock frequency and the utilization percentage of each core. In this paper, we only consider an asymmetric single-ISA CPU, however including other computational units (like GPUs and DSPs) would only impact the number of configuration parameters of the platform and require different types of workload. 

To that extent, we evaluate three system identification methods with respect to their prediction accuracy: a linear state-space identification approach using polynomial regressors, a NARX neural network approach and a recurrent neural network approach configured in an FIR model structure. We evaluate the three methods on an Odroid-XU4 board featuring an Exynos 5422 SoC and perform a set of experiments to evaluate their prediction accuracy using a 1-hour and a 6-hour dataset. The Exynos 5422 SoC is composed of a big.LITTLE octa-core mobile processor combining a Cortex-A15 and Cortex-A7 quad-core. We acknowledge the Odroid XU4 is a few years old platform at the time of writing this paper, however we consider that the proposed approach, with its benefits and drawbacks, can be applied to other modern SoC platforms. 

\section{System Identification and Selection of Methods}

System identification \cite{lennart} is a field which deals with  creating mathematical models of dynamical systems through statistical and machine learning approaches. 

Several works have utilized neural networks for thermal modeling and have been proposed in the past. Some of them \cite{4487059,paci2007exploring} propose white-box approaches that are based on the theoretical equation that governs the power and heat dissipation of a processor. These implement a bottom-up technique, where the thermal model is based on the layout of the SoC, the conditions of the external environment and the conductive properties of materials. These approaches simulate the thermal dissipation directly at chip-level with some level of abstraction. This type of modeling relies heavily on the accuracy of the technical parameters and how much detail is lost through abstractions and simplifications. 

Another approaches \cite{liu2009chip,7517635} use the thermal-electrical analogy, in which the chip is broken down into small parts; each part is represented as a combination of current sources, resistors and capacitors. A common tool for these is HotSpot \cite{1650228}. These approaches also rely heavily on knowledge about the characteristics as well as the location of components within the chip.

Several researchers applied gray-box identification approaches to model the thermal characteristics of a processor. Beneventi et al. \cite{6381401} propose an approach where a multi-core processor is modeled as a thermal-electrical circuit. In their approach, the processor is divided into blocks that correspond to each core and the section of the copper heat spreader directly above each core. The parameters of the model are then optimized using an Output-Error approach. A similar approach was proposed by Aguia et al. \cite{5456979}. They suggested an implementation where the cores of a multi-core processor and the cache memory are represented as blocks in a thermal-electrical-equivalent circuit. The subspace identification method, N4SID, is then applied to find the optimal parameters for the model. Another approach that utilizes a state-space identification method has been proposed in \cite{liu2014compact}. Here, the researchers deploy a piece-wise linear subspace identification method that estimates a linear model for each temperature range. Shetu et al. \cite{7490925}, however, suggest a different approach with a polynomial model for approximating the temperature of a CPU. In their study, a thermal model is constructed by creating polynomials based on the size and intensity of the workload. 

Several black-box approaches based on neural networks have been proposed.  Vincenzi et al. \cite{5993628} and Sridhar et al. \cite{6106739} predict thermal dynamics of an integrated circuit using ARX linear neural networks. These approaches were shown to be effective at simulating heat flow in three-dimensional and highly granular, integrated circuits. Zhang et al. \cite{zhang_machine_2018} use a feed-forward neural network to simulate the heat dissipation in processors
By comparing the performance of a Gaussian process model, a neural network model and a linear regression model the researchers showed that the neural network model outperformed the linear model in terms of prediction accuracy by 30\%, but was approximately three times more computationally expensive. The Gaussian process model also showed good prediction accuracy, at the expense of twice the computational overhead of the neural network model.

Pérez et al. \cite{perez} compared recurrent and feed-forward neural network structures for thermal prediction of immersive cooling computer systems. The core frequency and processor utilization measurements from the past minute were used for temperature predictions. 

Differently from the previous approaches which rely on power measurements to predict the temperature of a processor, work has been done on predicting the power dissipation of a processor. Walker et al. \cite{walker2} predict the power consumption of a multi-core processor by utilizing core frequencies, core voltages and event counters (e.g., cycle counter, bus and cache accesses) to train a linear regression model. Zhang et al. \cite{zhang} built a linear regression model based on data collected from a CPU, where they utilized the idle states and idle time of each core. In addition, Balsini et al. \cite{balsini} deploy a genetic algorithm to find the optimal parameters for a function that represents the theoretical relationship between power dissipation and quantities such as the core voltage and clock frequency. 

In reality, as most models are constructed based on some knowledge and observations of a system, their corresponding modeling approaches can be viewed as being gray-box approaches to some degree \cite{janczak}. Most of the white-box and gray-box approaches utilize power as an input variable or \textit{regressor}. When the thermal properties and the blueprint are not directly available for the ARM CPUs, a white-box approach is not suitable. Many gray-box approaches also relied on the close-to-linear relationship between temperature and power. This also makes these approaches less appealing when the objective of this work is to perform modeling based only on measurable processor state variables, like frequency and processor utilization. However, some previous works exploit the theoretical relationship between frequency, voltage and utilization to estimate the power dissipation of a processor \cite{walker2} \cite{balsini}. Therefore, combining such an approach with a linear model identification technique, such as the N4SID method suggested in \cite{5456979}, was selected as an approach to be evaluated in this work.  

Other approaches that have produced promising results are neural network-based approaches \cite{zhang_machine_2018,6106739}. A neural network in an ARX structure, could through the addition of a nonlinear hidden layer, learn to replicate the nonlinear dynamics of the heterogeneous processor. This would create a Hammerstein type of NARX model. As the dynamics of a heterogeneous processor is rather deterministic and the noise component in measurements can be expected to be rather low, an ARX-based model was also selected as an approach to be evaluated in this work. 

Recurrent neural network approaches have not seen much attention in applications related to thermal modeling of computing systems. However, the approach in \cite{perez}, where where an RNN model is trained in an FIR structure, showed promising results. We therefore selected such RNN-based model as an approach to be evaluated.

The above approaches were not applied to create prediction models for thermal dissipation of heterogeneous SoCs. Thus in this paper we provide two contributions: a) we evaluate some of the proposed methods in the context of thermal dissipation and b) we propose a new approach, polynomial N4SID, as a combination of two existing methods.

\section{Evaluated methods}
Based on the surveyed literature three methods have been selected for comparison. They will be described in the following:  the first is a polynomial extension to N4SID, which we denote hereafter as Polynomial N4SID, a nonlinear state-space model structure using nonlinear regressors. The second is a NARX approach, where a neural network is recursively trained to predict the temperature. The third approach is an FIR model structure that utilizes an RNN layer to predict the thermal dissipation.
The performance of these three modeling approaches has been assessed for two different lengths of training data: 1 hour and 6 hours.  The error of each model has been measured using Mean Squared Error (MSE) as the metric. All three methods start with 10 regressors, i.e., the two cluster frequencies and the utilization of each of the eight cores.

\subsection{Polynomial N4SID}
The first model structure is a parametric approach based on the state-space identification method N4SID to estimate a linear state-space model.  There is a direct relationship between the power dissipation of a processor and its thermal dissipation. This relationship could, therefore, be exploited to construct a linear model of the system. This type of approach has been suggested in both \cite{6381401} and \cite{5456979}. In this work, however, the objective is to compare modeling approaches that can predict the temperature based on the clock frequency and the utilization percentage of each core. The power consumption has a nonlinear relationship with the core frequency, the core voltage and the core utilization. While the dynamic power dissipation is linearly dependent on the core utilization, the core utilization cannot, on its own, be used to describe it, as it is also dependent on the core frequency and voltage. Therefore, a non-linearity relation had to be introduced to approximate the power dissipation, in the form of new nonlinear regressors as polynomial combinations of the core frequency and core utilization. 

\newcommand{\appropto}{\mathrel{\vcenter{
  \offinterlineskip\halign{\hfil$##$\cr
    \propto\cr\noalign{\kern2pt}\sim\cr\noalign{\kern-2pt}}}}}

In our work, we approximate the relation between the voltage and the core frequency as $V \appropto \sqrt{f} $. The dynamic part of the power consumption is expressed as $P_{dyn} \appropto f^{2}$, while the static part of the power consumption was estimated to be approximately proportional to $f^{1.5}$. In this scenario, the core utilization is expected to be directly proportional to the dynamic power consumption. 

Using these approximate relationships as a basis, the polynomials were created as the product of the core utilization to a power of 0 or 1 and the core frequency to a power of between 1 and 3 in increments of 0.5. This was performed for each core and resulted in 58 new nonlinear regressors with a total of 68 regressors, including the original 10.

The N4SID algorithm does not have many parameters that can be tuned. However, the model order can be viewed as a hyperparameter. In this implementation, the selection of nonlinear regressors can also be considered as hyperparameters. Optimization of the utilized regressors was performed using correlation analysis and grid search. 
 
\begin{figure}
\centering
  \includegraphics[width=\linewidth]{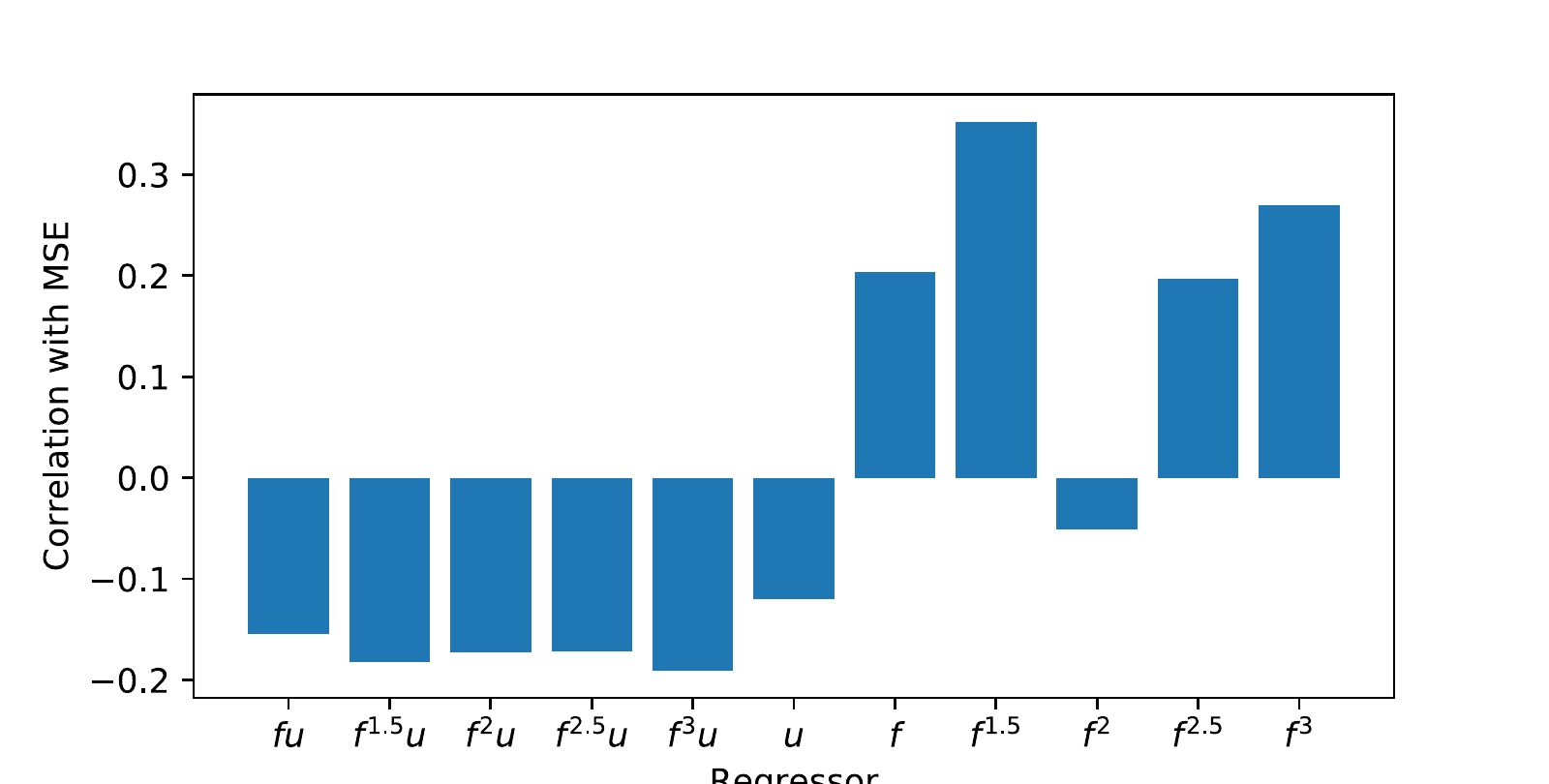}
  \caption{Correlation between regressor and MSE.}
  \label{fig:corr}
\end{figure}

A randomized search was performed over 500 iterations on values measured over a one hour workload execution time, as describe in Section 4.
In each iteration, three random combinations of core frequency to a power between 1 and 3 and core utilization to a power of 0 or 1 were selected. The combinations were then applied to the regressors belonging to each core to create the new regressors. At the end of each iteration, the average mean square error (MSE) was measured. Using the results, a pair-wise correlation analysis was performed to detect overall contribution of each regressor to the error. Figure \ref{fig:corr} shows that most of the regressors with only a single frequency component showed a positive correlation. That is, they increased the error when they were utilized. Those that showed a negative correlation produced a decrease in the error when they were utilized. The regressors with a positive correlation were therefore removed from the regressor set.

Grid search and cross-validation were performed as an additional reduction step. During the grid search, the model order was set to 5 for all iterations. This was implemented to reduce computational time. The model order that produces the best performance was, however, expected to be higher than 5. An assumption was made, though, that a fifth-order model would be representative enough for this hyperparameter validation step. All permutations of the remaining regressors were tested and the best regressor configuration was saved. The best regressor set is shown in \eqref{eq:n4sidregessors}, where $f$ is core frequency, $u$ is core utilization and $i$ indicates the number of cores. 

\begin{equation}
\label{eq:n4sidregessors}
U_{nl} = [f^{1.5}u_i, f^{2}u_i, f^{3}u_i, u_i, f^2] , i=1..8
\end{equation}

\begin{wrapfigure}{r}{0.7\textwidth}
\centering
  \includegraphics[width=\linewidth]{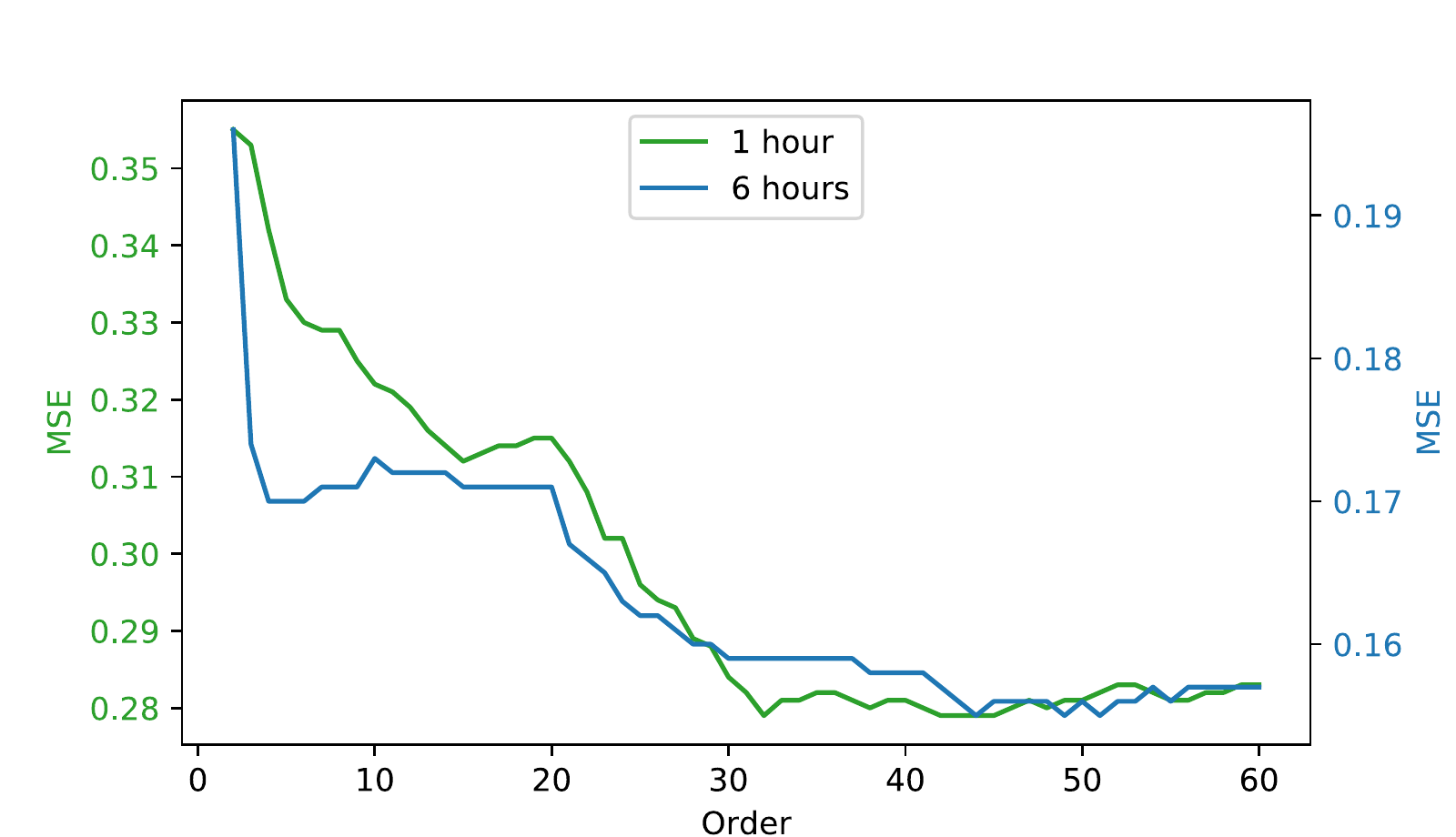}
  \caption{Validation error and model order.}
  \label{fig:orders}
\end{wrapfigure}

The final number of regressors utilized in this approach is 34. Furthermore, these regressors were selected for implementation for both 1-hour and 6-hour block lengths.  The average validation error was measured for orders between 2 and 60. Figure \ref{fig:orders} shows the model performance for each order. We presented this approach in detail in \cite{JoelIteqs2021}.

\subsection{Hammerstein-NARX}
\begin{figure}
\centering
  \includegraphics[width=0.7\linewidth]{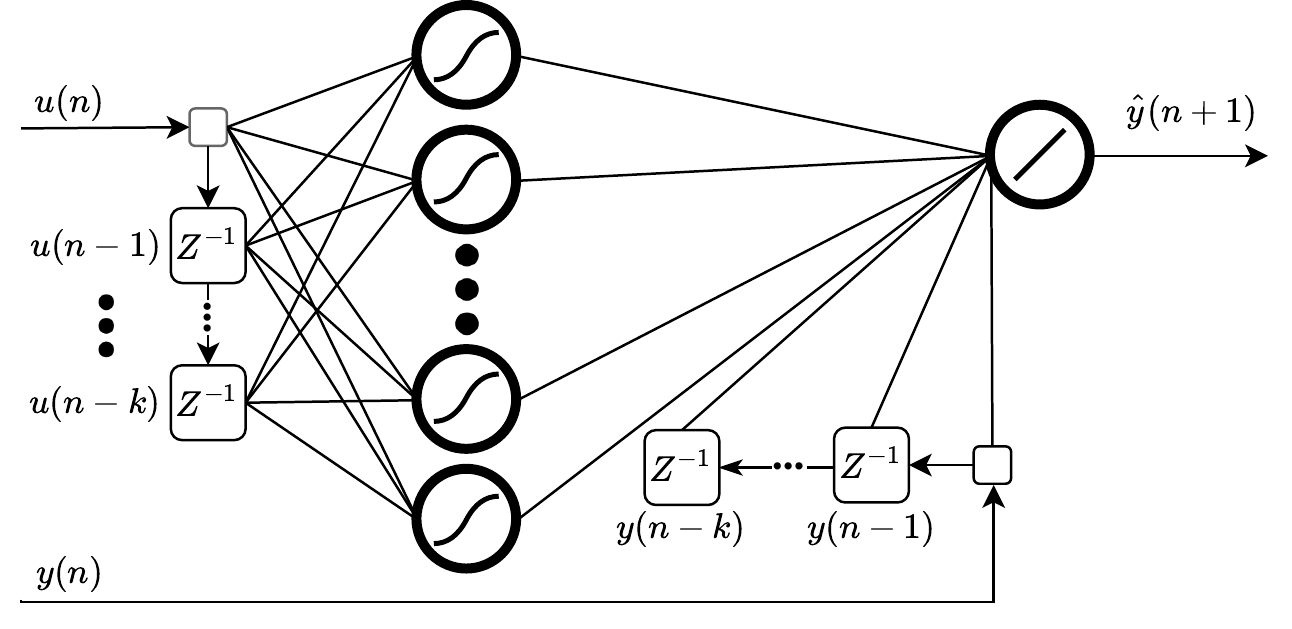}
  \caption{Offline Hammerstein-NARX structure used for training.}
  \label{fig:narxopen}
\end{figure}

The second model structure chosen was an NARX approach implemented as an artificial neural network. As shown by Zhang et al. \cite{zhang_machine_2018} and Sridhar et al. \cite{6106739}, a neural network can be trained to predict the temperature at the next time step based on previous temperature values and some exogenous inputs that affect the temperature. The two approaches were in this implementation combined to create a Hammerstein-NARX structure. In this approach, a network with one hidden nonlinear layer and one linear output layer has been constructed. The inputs are the 10 regressors and their respective values shifted back in time $n_x$ time steps. The nonlinear layer uses a sigmoid activation function to approximate the nonlinearity of the system. The output layer is a linear function that produces a weighted sum of the values that are produced by the nonlinear layer. The output from the linear output layer is fed back to itself for the past $n_y$ time steps. Figure \ref{fig:narxopen} shows how the network was structured during training.

\begin{figure}
\centering
  \includegraphics[width=0.7\linewidth]{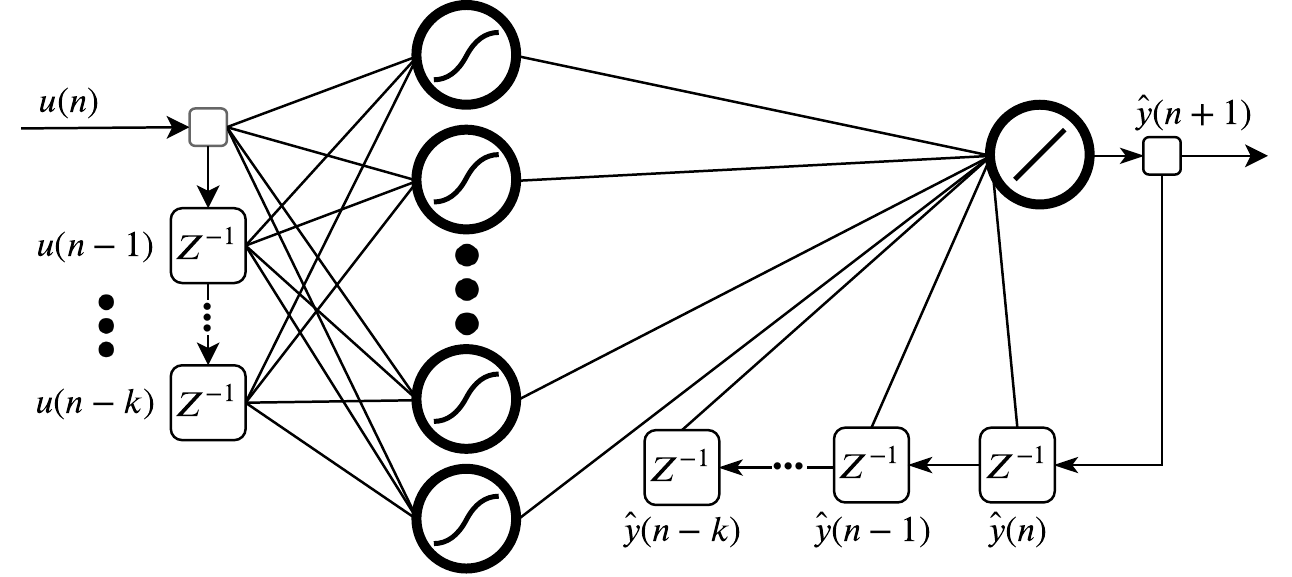}
  \caption{Online Hammerstein-NARX structure used to produce predictions.}
  \label{fig:narxclosed}
\end{figure}

The network is trained in an offline configuration. This was chosen since the online configuration suffered from the vanishing gradient problem during training. In an offline configuration, there is no recurrence in the network. Thus, the vanishing gradient is not an issue. 
Early stopping on the validation performance was implemented as well. The training was stopped when the error on the validation set started to increase. When the training of the network was finalized, the model structure was closed to produce the online layout shown in Figure \ref{fig:narxclosed}. Using this structure, the network can generate predictions of future values of the temperature without relying on actual temperature measurements as inputs.

A few hyperparameters for this approach were selected based on the network structures suggested in \cite{zhang_machine_2018} and \cite{6106739}, as well as some empirical experience. The activation function was selected to be a sigmoid function. Additionally, only a shallow structure with one hidden layer was tested. The selected optimization algorithm, Levenberg–Marquardt, was also not changed and its associated parameters were kept as the default for the \textit{trainlm} function in Matlab's Deep Learning Toolbox. The Levenberg–Marquardt optimization algorithm was chosen since it was the only algorithm that could successfully converge to a solution during training on the offline configuration.

\begin{figure}[h]
\centering
  \includegraphics[width=.8\linewidth]{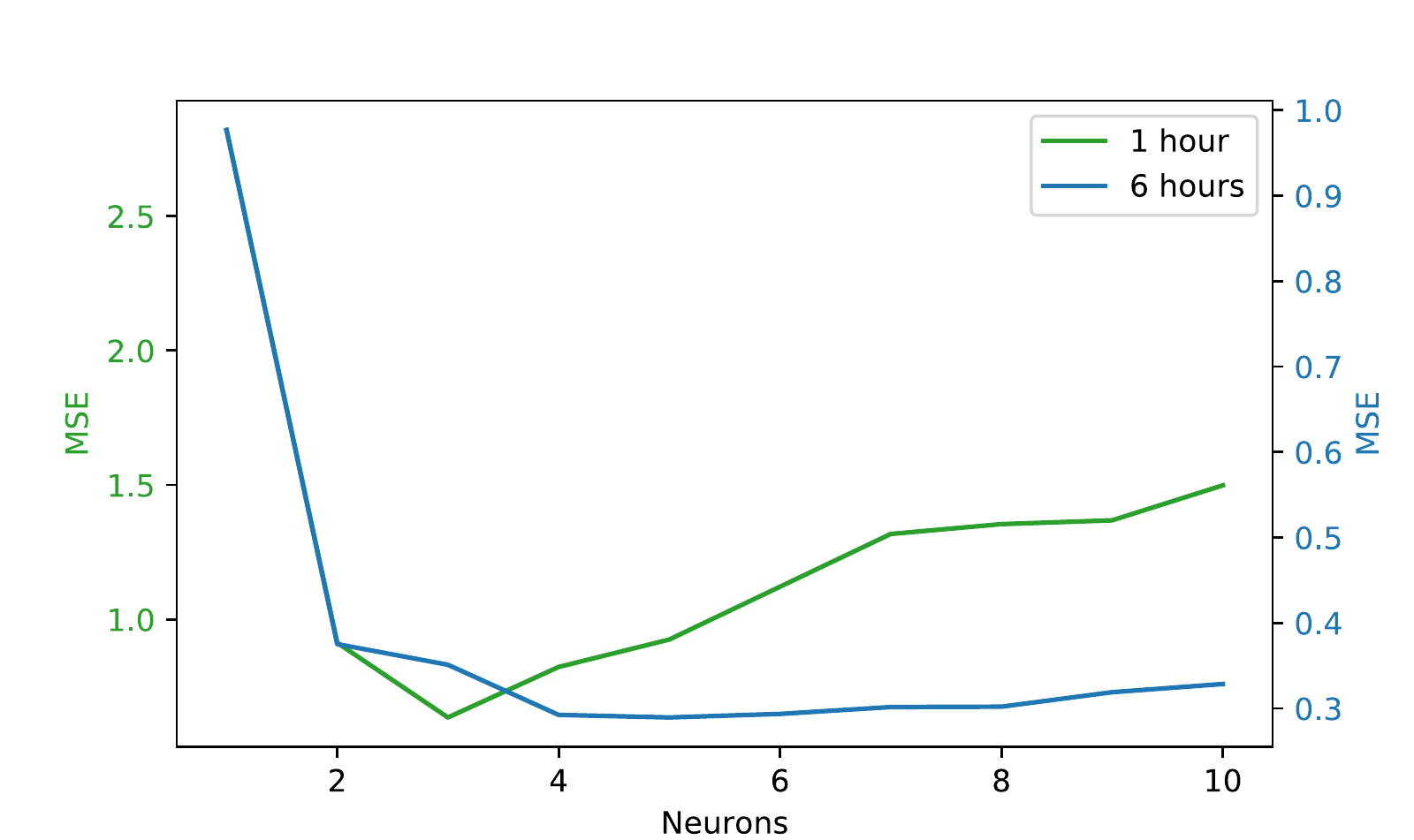}
  \caption{Validation error per size of the hidden nonlinear layer.}
  \label{fig:layers}
\end{figure}

In Figure \ref{fig:layers}, the validation performance for different layer sizes is shown. For the 1-hour block length, 3 neurons in the hidden layer produced the best validation performance on average. When training the model structure using 6 hours of data, 5 neurons yielded the lowest average prediction error.

\subsection{FIR-RNN}
The final model structure that was assessed is based on a recurrent neural network. This structure has one recurrent layer followed by a single linear layer. This is based on an FIR structure, where the output is predicted solely based on $n_x$ previous inputs. 

\begin{figure}
\centering
  \includegraphics[width=0.9\linewidth]{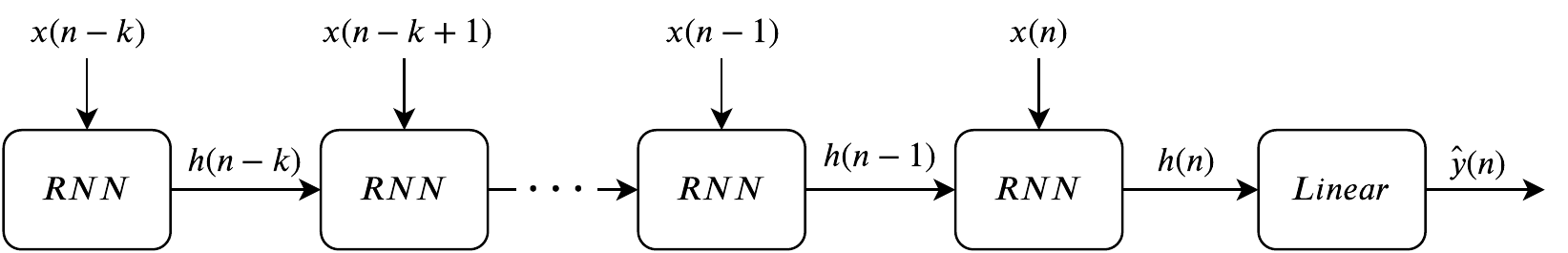}
  \caption{FIR-RNN model structure.}
  \label{fig:rnn_fir}
\end{figure}

This modeling approach is based on the structure utilized by Pérez et al. \cite{perez}. They found that a shallow structure with either GRU or LSTM layers produced the best performance in their immersive cooling experiment. A similar approach is therefore implemented, as shown in Figure \ref{fig:rnn_fir}. A single layer of RNN neurons is followed by a single linear layer. Each time step, the RNN layer takes the input vector $x$, which corresponds to the 10 original regressors, and passes it through the neurons to produce a vector of nonlinear states $h$ that is passed to the next time step. This is performed until the current time step is reached. The hidden state vector $h$ is then passed through a linear function to determine the prediction $\hat{y}$. The nonlinear function that is applied inside each recurrent unit differs depending on whether it is a GRU unit or an LSTM unit and on the activation function that is utilized. Early stopping on the validation set has also been utilized for this approach. 

The hyperparameters that were selected empirically were the optimizer and activation function utilized in the RNN nodes. Pérez et al. \cite{perez} utilize the Nesterov Adaptive Momentum (Nadam) optimizer and a \textit{tanh} activation function. Thus, these parameters were selected in this implementation, as well.

The first hyperparameter that was assessed was the number of time steps for the input that had to be considered. Since this approach is not recursive, many time steps have to be included to capture the response of the system. To estimate the settling time of the system, a step response was measured by going from 0 to 100\% utilization on all cores when the Odroid board was configured to run at 1800 MHz and 1500 MHz for the big and little cluster, respectively. Measurements shows that it takes approximately 100 seconds for the system to settle. Therefore, it can be concluded that an FIR model would need the input values for the past 100 seconds to be able to simulate the dynamics of the system accurately.

The sample rate and the number of samples were tested through grid search and cross-validation. Three other hyperparameters were also tested in conjunction: the unit type (LSTM or GRU), the number of units and the batch size. For both the 1-hour and 6-hour block lengths, a sample length of 50 samples spread out logarithmically between 0 and 100 seconds, performed the best. The GRU unit also outperformed the LSTM unit using both block lengths. A batch size of 1 and a unit size of 10 was found to be the optimal values for the 1-hour block length. Using the longer block length, a batch size of 4 and a unit size of 18 generated the lowest average validation error.

\section{Experimental setup}
\label{section:CaseStudy}

\begin{figure}
\centerline{\includegraphics[width=1\textwidth]{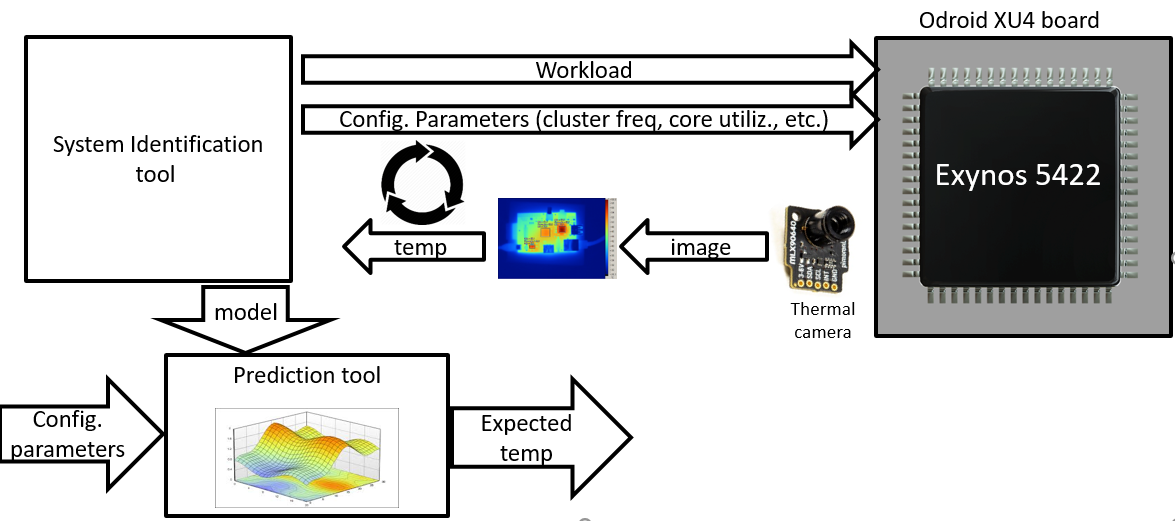}}
\caption{Overview of the experimental setup}
\label{overview}
\end{figure}

For this study, a desktop experiment setup to benchmark and measure the temperature of a heterogeneous processor was created. The experimental setup  in Figure \ref{overview} was used to generate and gather data in this study. 

\textbf{System Under Test}. In this case the system under test was an Odroid XU4 Exynos 5422 board - a single-board computer allows the control of the frequency on a per cluster basis between $200$ MHz and $2000$ MHz for the big cluster and $200$ MHz and $1500$ MHz for this little cluster. The operating frequency cannot be controlled independently for each core inside the clusters. The voltage levels can also be set for each cluster. However, in the Linux operating system for this platform, these are set to static values for each operating frequency by the kernel. The operating voltage levels are, therefore, not considered as a variable in the implementations in this work. 

The Odroid board has been configured to trigger a thermal throttle when the core temperature for the big cores reaches 90$^{\circ}$C. This means that the processor's frequency governors will reduce the maximum available frequency when the temperature is reached to prevent the processor from overheating.

\textbf{Experimental workload} utilized in this experiment was an RGB-YUV image conversion. This image conversion was chosen as the workload because it is a highly parallel workload that can be distributed to many cores.

A custom-built stress configures the platform by setting the utilization of each core, the frequency of each cluster, and the amount of time for the execution of the workload. Inside the application, a thread for each core in the system is created. Each core thread runs its assigned workload independently from the other cores. 

The cluster frequencies are controlled using the \textit{Performance} frequency governor. The used application does not adjust the frequencies directly, it sets the maximum allowed frequency and the frequency governor then adjusts the frequency accordingly. 

The workload is in this work constant. Thus, the total number of possible configurations can be calculated using Equation \eqref{eq:configurations}, where $U$ is the number of utilization levels, $C$ is the number of cores, $f_b$ is the clock frequency of the big cluster and $f_l$ is the frequency of the little cluster. 

\begin{equation}
\label{eq:configurations}
N_c = U^C f_b f_l
\end{equation}

Each core has five different utilization levels, and the big and little clusters have ten and six discrete clock frequency levels, respectively. For the implementation in this work, this yields a total of approximately 23 million possible configurations of the heterogeneous SoC. 

\textbf{Thermal measurements}
Due to the absence of a core temperature sensor for each core on the Odroid-XU4, a Melexis MLX90640 thermal camera was used. The camera, with a resolution of 32x24 pixels, has been mounted close to the SoC of the Odroid-XU4 in order to obtain a more accurate reading of the temperature across the surface of the SoC. The camera sensor has a temperature range of 40$^{\circ}$C to 300$^{\circ}$C and an accuracy of approximately $\pm 1 ^{\circ}$C. 

\textbf{Cooling}
Due to the use of the thermal camera, the heat sink the SoC was removed and the external cooling was provided via a fan as direct cooling which allowed the big cluster to be able to run at up to $1900$ MHz. For this work, the fan is constantly running at 100\% speed and the environment temperature has been kept constant at around 21 degrees.

\textbf{Data collection}
A Raspberry Pi has been deployed as the control and data collection unit. It controls the experimental workloads and captures the thermal response. The data from the temperature sensor and the Odroid board were sampled 32 times per second. This sample rate was selected since it is the maximum sample rate for the thermal sensor.  The data set for model selection was created by executing a sequence of randomly selected configurations of the Odroid board using the stress application mentioned above. The configuration of the board was changed after a random amount of time in the range of 10 to 60 seconds. Both the selection of configuration parameters (cluster frequencies and core utilization) and execution period followed a uniform distribution. Throughout the experiment, the ambient temperature was kept steady at 21$^{\circ} $C.

\section{Model Selection and Evaluation}

\begin{figure}[t]
\centering
  \includegraphics[width=0.66\linewidth]{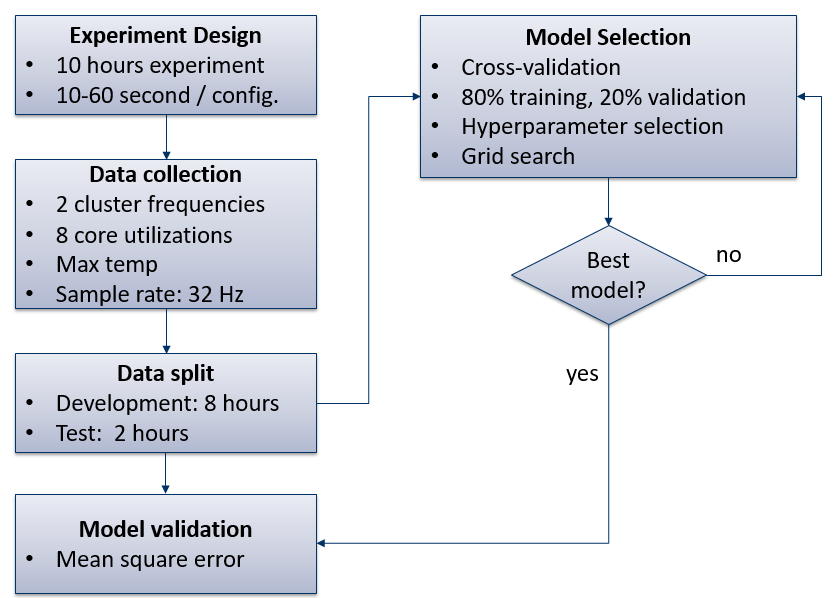}
  \caption{Flowchart of system identification procedure.}
  \label{fig:flowchart}
\end{figure}

The performance of the three selected modeling approaches has been assessed for two different lengths of training data: 1 hour and 6 hours. The error of each model has been measured using MSE as the metric. A flowchart of the entire model identification methodology is shown in Figure \ref{fig:flowchart}.

The previously collected data set was divided into two sets, a development set and a test set. The first 79\% of the data became the development set. This is the portion of the data that the models were trained on and the models' hyperparameters were evaluated with. The last 20\% of the data were chosen as the test set. This is the data set that the final prediction error was assessed upon and was not utilized for model training and selection. A small set of data corresponding to 1\% of the total data, lodged between the development and test sets, is omitted to ensure that there is no interference between the development set and the test set. Furthermore, the same data split was utilized for all three modeling approaches.

\textbf{1-hour performance. }
Using the hyperparameters and model structures described in the previous section, the models were validated through 10-fold cross-validation. Table \ref{tab:1hour-test} shows the result for the model on the 1-hour block length. 

\begin{table}
     \scriptsize
    \setlength\tabcolsep{3pt}
    \centering
    \captionsetup{justification=centering}
    \caption{MSE for the implemented approaches trained with 1 hour of data.}
    \label{tab:1hour-test}
    \begin{tabular}{|l|c|c|c|c|c|c|c|c|c|c|c|}   
    \hline
       & \multicolumn{10}{|c|}{\textbf{Folds}} & \\ \hline
        \textbf{Method} &1&2&3&4&5&6&7&8&9&10&\textbf{Avg} \\  \hline
        \textbf{Polynomial N4SID} &0.16	&0.15	&0.15	&0.16	&0.16	&0.14	&0.16	&0.16	&0.17	&0.14&\textbf{0.16}  \\ \hline
        \textbf{Hammerstein-NARX} & 0.53&1.28&0.61&0.74&0.74&0.55&0.65&0.85&0.79&0.54& \textbf{0.73 }\\ \hline
        \textbf{FIR-RNN} & 2.28&2.14&1.42&1.44&1.05&2.12&1.60&1.30&2.77&0.80 & \textbf{1.69}  \\ \hline
    \end{tabular}
\end{table}

Table \ref{tab:1hour-test} shows that the Polynomial N4SID approach showed the lowest average MSE. It can also be noted that the N4SID based approach has, by far, the lowest variance, with a standard deviation of just 0.01. The other two approaches had significantly worse performance on the test data.

\begin{figure}[h]
\centering
  \includegraphics[width=\linewidth]{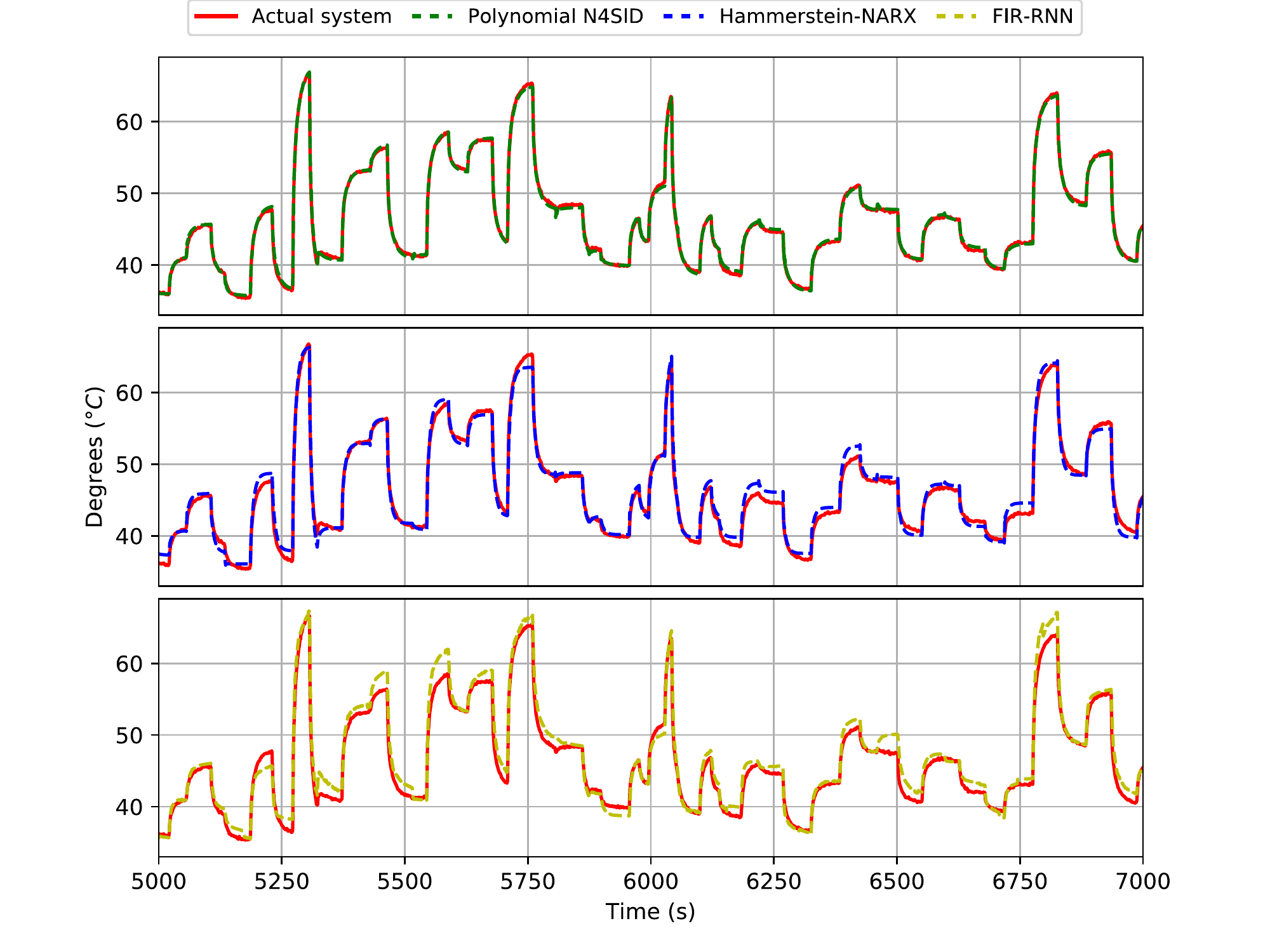}
  \caption{1-hour model predictions on the last 2000 seconds of the test data.}
  \label{fig:1hourplot}
\end{figure}

Figure \ref{fig:1hourplot} shows the models' performance on the test set when trained on the seventh fold. This fold is selected since it is the fold that is the closest to the average for all three approaches. The configuration parameters of the board were randomly changed every 10 to 60 seconds. This means that approximately 57 different board configurations were utilized in the 2000 second window.

Looking at the above figure, it can be seen that the Polynomial N4SID model produced a good approximation of the true measured temperature. The other two models produced less desirable results, but they still yielded a decent approximation of the true temperature. Furthermore, the Polynomial N4SID model does not appear to have any particular problem areas or specific configurations that it struggles with. The other two models and especially the FIR-RNN show varying performance in regards to the different board configurations.

The average training time, average prediction time and the number of parameters were also measured for the three model structures. Table \ref{tab:1hour-comp} shows that the N4SID-based model structure has the lowest training and prediction time. However, it is closely followed by the Hammerstein-NARX model structure. The FIR-RNN model takes the longest both to train and to make predictions. The training time is especially significant as it is about 100 times that of the other two approaches. 

\begin{table}[h]
     \scriptsize
    \setlength\tabcolsep{3pt}
    \centering
    \captionsetup{justification=centering}
    \caption{Average training time, average prediction time and number of parameter for the 1-hour models.}
    \label{tab:1hour-comp}
    \begin{tabular}{|l|c|c|c|}   
    \hline
        \textbf{Method}&Training time (s)&Prediction time (s)&Number of parameters \\  \hline
        \textbf{Polynomial N4SID} &6	&0.25	&2144  \\ \hline        \textbf{Hammerstein-NARX} & 7&0.558&347\\ \hline
        \textbf{FIR-RNN} & 987&4.9&671\\ \hline
    \end{tabular}
\end{table}

\textbf{6-hour performance.}
 The same procedure was utilized for the 6-hour block length. The models were validated through 4-fold cross-validation. Table \ref{tab:6hour-test} shows the result for the model when trained with 6 hours of data.

\begin{table}
     \scriptsize
    \setlength\tabcolsep{3pt}
    \centering
    \captionsetup{justification=centering}
    \caption{MSE for the implemented approaches trained with 6 hours of data.}
    \label{tab:6hour-test}
    \begin{tabular}{|l|c|c|c|c|c|}   
    \hline
       & \multicolumn{4}{|c|}{\textbf{Folds}} & \\ \hline
        \textbf{Method} &1&2&3&4&\textbf{Avg} \\  \hline
        \textbf{Polynomial N4SID} & 0.11 & 0.11 & 0.11 & 0.11 &  \textbf{0.11}  \\ \hline
        \textbf{Hammerstein-NARX} &0.26&0.25&0.28&0.28&\textbf{0.27 }\\ \hline
        \textbf{FIR-RNN} & 0.24 &0.21 &0.18 &0.19 & \textbf{0.21}  \\ \hline
    \end{tabular}
\end{table}

\begin{figure}[h]
\centering
  \includegraphics[width=\linewidth]{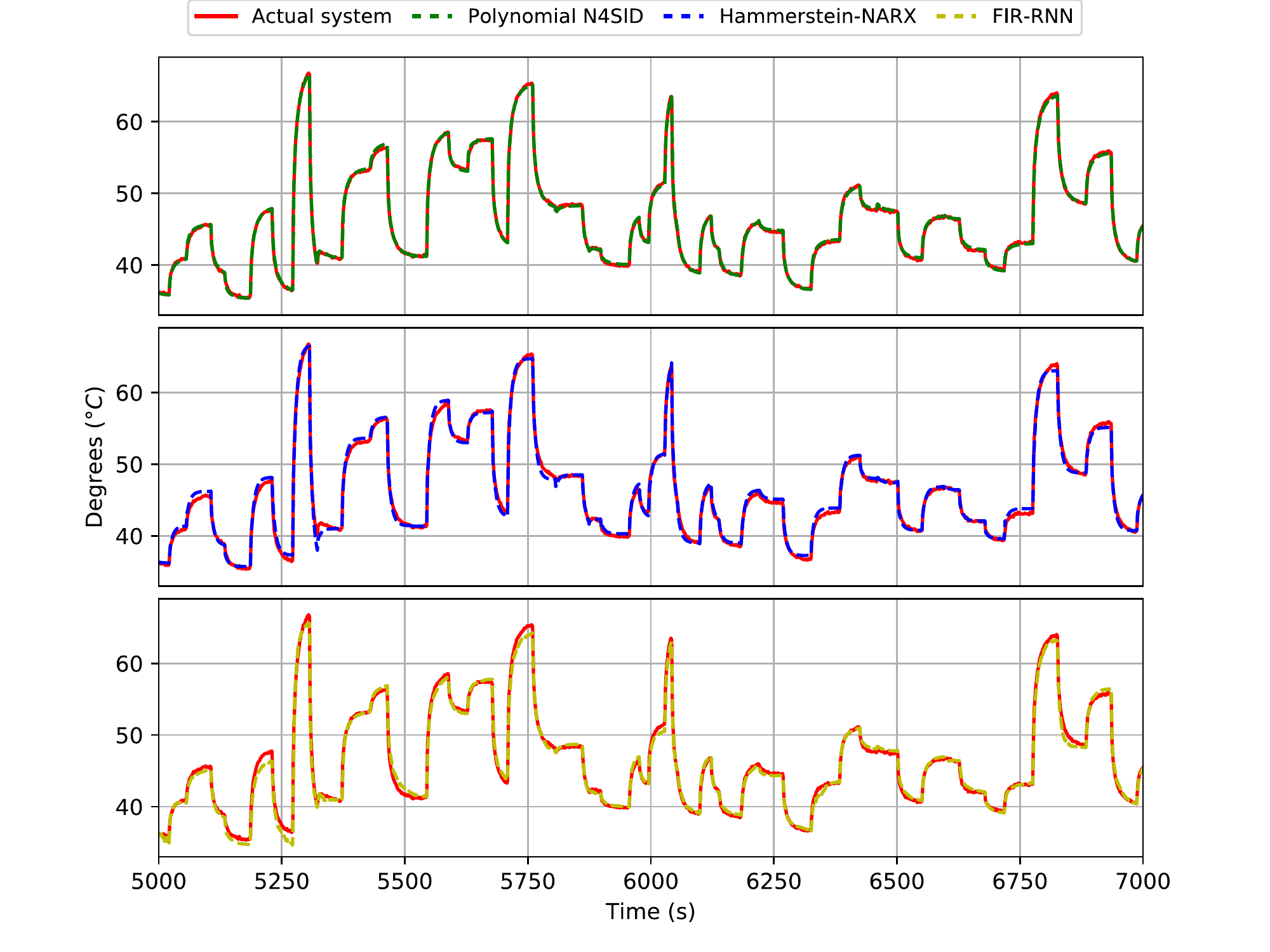}
  \caption{6-hour model predictions on the last 2000 seconds of the test data.}
  \label{fig:6hourplot}
\end{figure}

The prediction error of the Polynomial N4SID model was reduced even further when trained with 6 hours of data. It improved by approximately 50\% compared to its 1-hour performance. The FIR-RNN model, however, has improved substantially. It yields a prediction MSE of 0.21 when trained with more data. The Hammerstein-NARX model did also improve compared to the 1-hour block length, but it did not see the same level of improvement as the recurrent FIR model. Figure \ref{fig:6hourplot} shows the three modeling approaches' performance on the final 2000 seconds of the test set when trained on the second fold.

The average training time, average prediction time and the number of parameters for the three model structures on 6 hours of training data is shown in Table \ref{tab:6hour-comp}. Just as for 1 hour of training data, the N4SID and NARX-based models have significantly lower training and prediction times. Interestingly, the Hammerstein-NARX model's prediction time only increased slightly and is more than twice as fast as the Polynomial N4SID model.

\begin{table}
        \scriptsize
    \setlength\tabcolsep{3pt}
    \centering
    \captionsetup{justification=centering}
    \caption{Average training time, average prediction time and number of parameter for the 1-hour models.}
    \label{tab:6hour-comp}
    \begin{tabular}{|l|c|c|c|}   
    \hline
        
        \textbf{Method}&Training time (s)&Prediction time (s)&Number of parameters \\  \hline
        \textbf{Polynomial N4SID} & 60	 & 1.34	  & 3354  \\ \hline        \textbf{Hammerstein-NARX} & 67   & 0.65   & 571  \\ \hline
        \textbf{FIR-RNN}          & 2580 & 10.5   & 1639  \\ \hline
    \end{tabular}
\end{table}

\section{Conclusion}

The results of this study show that several types of modeling approaches can be utilized to predict the temperature dissipation of a heterogeneous SoC. However, Polynomial N4SID outperforms the others in its ability to learn the dynamics of a system from a limited amount of data and to predict with higher accuracy the thermal dissipation. We consider that this is because the non-linear regressors are able to  better estimate the quadratic relationship between frequency and power consumption.

Future work is intended to address some of the limitations of the current study, for instance to consider several types of SoCs and workloads, to take into account the ambient temperature and humidity, and to investigate the accuracy of our models in a real-world setup without removing the heat sink and providing constant active cooling.

\end{document}